\RequirePackage[svgnames]{xcolor}

\documentclass[11pt,letterpaper]{mystyle}

\usepackage[all]{hypcap}
\usepackage[svgnames]{xcolor}
\usepackage[comma,authoryear,compress]{natbib}
\usepackage{hyperref}[citecolor=lightblue]

\hypersetup{
    colorlinks = true,
    citecolor = {YaleBlue},
}

\usepackage{algorithm}
\usepackage{algorithmicx}
\usepackage{algpseudocode}
\usepackage{microtype}
\usepackage{graphicx}
\expandafter\def\csname ver@subfig.sty\endcsname{}
\usepackage{booktabs} %
\usepackage{float}
\usepackage{bigstrut}

\usepackage{amsmath}
\usepackage{amssymb}
\usepackage{mathtools}
\usepackage{amsthm}
\usepackage{mathrsfs}
\usepackage{nicefrac}
\usepackage{dsfont}
\usepackage{enumitem}
\usepackage{subcaption}
\usepackage{graphicx,subfig}
\usepackage{bxcoloremoji}

\usepackage{float}

\usepackage[utf8]{inputenc} %
\usepackage[T1]{fontenc}    %
\usepackage{hyperref}       %
\usepackage{url}            %
\usepackage{booktabs}       %
\usepackage{amsfonts}       %
\usepackage{nicefrac}       %
\usepackage{microtype}      %
\usepackage{graphicx}
\usepackage{subcaption} 
\usepackage{fdsymbol}
\usepackage{wrapfig}
\usepackage{lipsum}
\usepackage{enumitem}
\usepackage{stackengine}
\usepackage[font=small,labelfont=bf]{caption}
\usepackage{color}
\usepackage{adjustbox}

\usepackage{rotating}
\usepackage{makecell}

\usepackage{amsmath}
\usepackage{mathtools}
\usepackage{amsthm}
\usepackage{xspace}
\usepackage[utf8]{inputenc} 
\usepackage[T1]{fontenc}    
\usepackage{amsfonts}       
\usepackage{nicefrac}       
\usepackage{microtype}      
\usepackage{xcolor}         
\usepackage{tikz}
\usepackage{booktabs}    
\usepackage{multirow}    
\usepackage{xcolor}      
\usepackage{graphicx}    
\usepackage{array}       
\usepackage{ulem}        
\usepackage{colortbl}
\usepackage{subcaption}
\usepackage{wrapfig}  
\usepackage{tabularx}
\usepackage[capitalize]{cleveref}

\newcommand{\x}{\mathbf{x}}

\definecolor{blanchedalmond}{rgb}{1.0, 0.92, 0.8}
\definecolor{carmine}{rgb}{0.59, 0.0, 0.09}
\definecolor{lightblue}{rgb}{0.22,0.45,0.70}%

\renewcommand{\mathbf}{\boldsymbol}

\makeatletter
\def\Ddots{\mathinner{\mkern1mu\raise\p@
\vbox{\kern7\p@\hbox{.}}\mkern2mu
\raise4\p@\hbox{.}\mkern2mu\raise7\p@\hbox{.}\mkern1mu}}
\makeatother

\definecolor{amaranth}{rgb}{0.9, 0.17, 0.31}
\definecolor{antiquebrass}{rgb}{0.8, 0.58, 0.46}
\definecolor{antiquefuchsia}{rgb}{0.57, 0.36, 0.51}
\definecolor{chromeyellow}{rgb}{0.31, 0.47, 0.26}

\newtcolorbox{AIbox}[2][]{aibox,title=#2,#1}
\definecolor{lightblue}{rgb}{0.22,0.45,0.70}%
\definecolor{Gray}{gray}{0.95}
\definecolor{Cornsilk}{rgb}{1.0, 0.97, 0.86}
\definecolor{hightlightyellow}{rgb}{1.0, 0.827, 0.278}

\usepackage{amsmath}

\usepackage[all]{hypcap}

\title{Embodied Arena: A Comprehensive, Unified, and Evolving Evaluation Platform for Embodied AI}

\runningtitle{Embodied Arena: A Comprehensive, Unified, and Evolving Evaluation Platform for Embodied AI}

\author{
\textsuperscript{$\alpha$}\textbf{Tianjin University}, \textsuperscript{$\beta$}\textbf{Huawei Noah's Ark Lab}, 
\textsuperscript{$\gamma$}\textbf{Shanghai Jiao Tong University},
\\
\textsuperscript{$\delta$}\textbf{Hong Kong University of Science and Technology (Guangzhou)},
\textsuperscript{$\lambda$}\textbf{Sun Yat-sen University}, \textsuperscript{$\rho$}\textbf{PengCheng Laboratory},
\textsuperscript{$\zeta$}\textbf{Tongji University}, 
\textsuperscript{$\epsilon$}\textbf{University College London},
\\
\textsuperscript{$\mu$}\textbf{Peking University}, 
\textsuperscript{$\nu$}\textbf{Tsinghua University}, 
\textsuperscript{$\sigma$}\textbf{Imperial College London},
\textsuperscript{$\xi$}\textbf{King's College London},
\textsuperscript{$\kappa$}\textbf{Institute of Computing Technology, Chinese Academy of Sciences},
 \textsuperscript{$\eta$}\textbf{University of Manchester},
 \textsuperscript{$\theta$}\textbf{Nanjing University}, 
 \textsuperscript{$\pi$}\textbf{TU Darmstadt}
 
\textit{Full author list in Contributions}
}

\begin{document}

\begin{abstract}
Embodied AI has shown great promise in empowering AI models to perceive, interact with, and ultimately change the physical world.
Parallel to the development of large foundation models, Embodied AI is largely falling behind.
Located at the center of Embodied AI, three essential challenges emerge and become even more stringent: (1) systematic understanding of the core capabilities needed for Embodied AI is missing in the community, making research lack of clear objectives; (2) despite the proposal of various benchmarks for Embodied AI, there is no unified and standardized evaluation system, leaving the cross-benchmark evaluation and comparison infeasible; (3) different from large language models (LLMs) powered by numerous web-scale data, automated and scalable acquisition methods for embodied data have not been well developed, which poses a critical bottleneck on the scaling of evaluation and training of Embodied AI models.
To break the three obstacles, this paper presents \textbf{Embodied Arena}, a comprehensive, unified, and evolving evaluation platform and leaderboards for Embodied AI.
First, Embodied Arena is established upon \textbf{a systematic embodied capability taxonomy} spanning three levels (i.e., perception, reasoning, task execution), seven core embodied capabilities, and 25 fine-grained dimensions.
This taxonomy is proposed by absorbing and refining the partial categories in prior works, which allows for unified evaluation and offers systematic objectives for frontier research.
Second, Embodied Arena closes the critical gap in standardized evaluation by introducing \textbf{a unified embodied evaluation system}.
The system is built upon a unified evaluation infrastructure supporting flexible integration of advanced benchmarks and models, which has covered 22 diverse benchmarks across three domains (2D/3D Embodied Q\&A, Navigation, and Task Planning) and 30+ advanced models from 20+ worldwide institutes.
Third, Embodied Arena is powered by \textbf{a novel LLM-driven automated generation pipeline} that ensures the scalability of embodied evaluation data and allows it to keep evolving for diversity and comprehensiveness.
Building upon the three major components, Embodied Arena addresses the three essential challenges correspondingly.
Moreover, Embodied Arena provides professional support for more advanced models and embodied benchmarks to join, along with frequent maintenance and updates.
Through comprehensive evaluation of the growing model population based on evolving evaluation data, Embodied Arena publishes three types of leaderboards (i.e., Embodied Q\&A, Embodied Navigation, Embodied Task Planning) with two orthogonal views (i.e., the benchmark view and the capability view), offering a real-time overview of the embodied capabilities of advanced models.
Especially, we present \textbf{nine findings} summarized from the evaluation results on the leaderboards of Embodied Arena.
This helps to establish clear research veins and pinpoint critical research problems, thereby driving forward progress in the field of Embodied AI.

\vspace{2mm}

\coloremojicode{1F3E0} \textbf{Website}: \href{https://embodied-arena.com}{https://embodied-arena.com}

\vspace{2mm}
\noindent\rule{\textwidth}{0.4pt}
\vspace{1mm}

\end{abstract}

\maketitle
\vspace{3mm}
\vspace{-4mm}
\section{Introduction}
\label{sec:intro}

On the road towards Artificial General Intelligence (AGI), Embodied AI or Embodied Intelligence, has emerged to be one of the most important research fields in recent years.
Complementary to the general understanding, reasoning, tool-use and problem-solving abilities of large foundation models~\citep{openai_gpt3.5,jaech2024openai,guo2025deepseek,team2025kimi,qwen3}, Embodied AI has shown the promise in building various physical agents that are capable of perceiving, interacting with, and ultimately changing the real world~\citep{brohan2023rt,DBLP:conf/rss/BrohanBCCDFGHHH23,DBLP:conf/icml/DriessXSLCIWTVY23,o2023open,ZhaoKLF23ACT,team2024octo,kim2024openvla,black2024pi_0,DBLP:journals/corr/abs-2504-16054}.
Notably, OpenVLA~\citep{kim2024openvla} scaled vision-language-action models to enable generalist robotic manipulation across diverse tasks and \texttt{$\pi_0$}~\citep{black2024pi_0} introduced efficient hierarchical planning that bridges high-level reasoning with low-level control, demonstrating the potential of multimodal foundation models in embodied scenarios.

Although notable results have been achieved by the works above, there is still a huge gap between the capabilities of existing embodied agents and complex, diverse real-world application scenarios, preventing the wide-range deployment of Embodied AI techniques.
Concurrent with the development of large foundation models, Embodied AI is largely falling behind. 
Specifically, there are \textbf{three critical challenges} that severely limit the advancement of Embodied AI research.
First, \textit{what are the core capabilities that a desired Embodied AI model needs?} 
The answer to this essential question remains unclear.
Most ongoing works attempt to push the frontier from concrete aspects such as embodied visual perception, embodied task planning, etc. The lack of anchoring from a systematic view makes it hard to better connect with related works and find an important research purpose.
Second, despite the proposal of various benchmarks for Embodied AI, each benchmarks differ a lot in aspects like data formats, evaluation metrics, target embodied capabilities, etc. This makes direct cross-benchmark evaluation and comparison infeasible. Hence, a unified and standardized evaluation system is urgently needed.
Third, as the success of large language models (LLMs) stems from scalable training from numerous web-scale data, sufficient and diverse embodied data is crucial to thorough evaluation and training of Embodied AI models.
However, most existing embodied data relies heavily on manual scenario construction, task design, and data collection, making scalability impossible.
Unfortunately, a scalable, automated acquisition method for embodied data is missing in the field of Embodied AI, which poses a bottleneck on the scaling of evaluation and training of Embodied AI models.

To break the three obstacles and pave the way for the advancement of Embodied AI research, this paper presents \textbf{Embodied Arena}, the first comprehensive evaluation platform and leaderboards for Embodied
AI. 
First of all, we propose a \textbf{Systematic Embodied Capability Taxonomy} spanning three incremental levels (i.e., perception,
reasoning, task execution), seven core embodied capabilities, and 25 fine-grained dimensions. 
This taxonomy is established by absorbing and refining the partial categories in prior works, which allows for unified evaluation while offering systematic targets for frontier research.
Based on the systematic embodied capability taxonomy, Embodied Arena then closes the critical gap in standardized evaluation by introducing a \textbf{Unified Embodied Evaluation System}.
The system is built upon a unified evaluation infrastructure supporting flexible integration of advanced benchmarks and models, which has covered 22 diverse benchmarks across three domains (2D/3D Embodied Q\&A, Navigation, and Task Planning) and 30+ advanced models from 20+ worldwide institutes.
Embodied Arena also provides professional support for more advanced models and embodied benchmarks to join, along with frequent maintenance and updates.
Moreover, Embodied Arena is powered by a novel \textbf{LLM-driven Automated Data Generation Approach} for Embodied AI.
By leveraging the general knowledge in LLMs, this approach automates the whole process of scenario construction, task design, and data collection. 
This automated generation pipeline ensures the scalability of embodied evaluation data and allows it to keep evolving for diversity and comprehensiveness.
Building upon these major components, Embodied Arena addresses the three essential challenges correspondingly.

Through comprehensive evaluation among the growing model population based on evolving evaluation data, Embodied Arena publishes three types of leaderboards with two orthogonal views, i.e., \textit{benchmark view} and \textit{capability view}.
The benchmark view presents the ranking of models on each benchmark, which is convenient for academic researchers to quote and compare with in their research works; while the capability view instead presents the ranking of models against each embodied capability in the systematic taxonomy, providing an up-to-date overview of the embodied capabilities of advanced models.
Through carefully summarizing the comprehensive evaluation results on the leaderboards, especially, we present \textbf{nine findings} from a range of important perspectives for useful insights, including the comparison between general multimodal models and embodied models, the limitations of existing benchmarks, the relationship among different embodied capabilities, the scaling law of embodied AI, etc.
The ultimate aim of Embodied Arena is to facilitate the establishment of clear research veins and help to identify critical research problems, thereby propelling research progress in the field of Embodied AI.

In the following, we first presents an overview of Embodied Arena and highlight the key features in Section~\ref{sec:overview}.
Then we introduce the systematic embodied capability taxonomy as well as a mapping from existing benchmarks to our taxonomy in Section~\ref{sec:taxonomy}.
Thereafter, we detail our unified embodied evaluation system in Section~\ref{sec:architecture}, followed by the LLM-driven automated generation pipeline for embodied evaluation data in Section~\ref{sec:generation}.
Moreover, we summarize nine major findings from our comprehensive evaluation results to provide useful insights that illuminate the current state and future directions of
Embodied AI research in Section~\ref{sec:insights}.
Finally, the conclusion is summarized in Section~\ref{sec:conclusion} and all authors are listed in Section~\ref{sec:contribution}.

\vspace{-1mm}

\section{Overview of Embodied Arena}
\label{sec:overview}

\begin{figure}[t]
    \centering
    \vspace{-0.2cm}
    \includegraphics[width=0.75\linewidth]{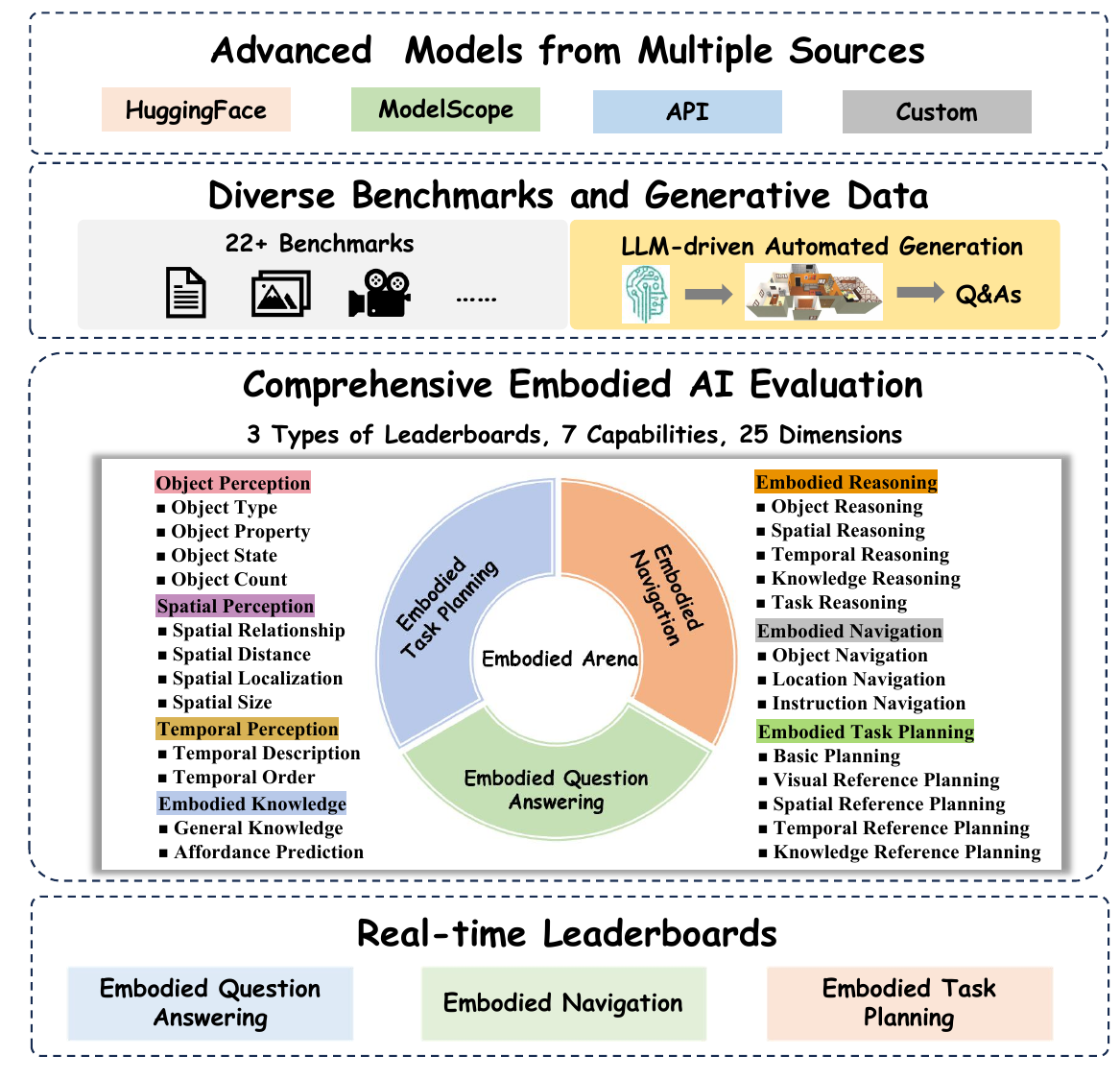}
    \vspace{-0.3cm}
    \caption{\textbf{A conceptual overview of Embodied Arena.} Embodied Arena provides a comprehensive evaluation for advanced models from multiple sources, based on diverse embodied benchmarks and LLM-driven generative data. The evaluation results span three types of embodied tasks (i.e., Embodied Q\&A, Embodied Navigation, Embodied Task Planning), against seven embodied capabilities with 25 fine-grained dimensions. Three types of leaderboards are summarized and presented for convenient and useful reference to both academia and industry.}
    \label{fig:Embodied Arena_overview}
\vspace{-0.2cm}
\end{figure}

Embodied Arena is a comprehensive, unified, and evolving evaluation platform and leaderboards for Embodied AI.
It features three types of core embodied tasks, a diverse range of high-quality benchmarks, an LLM-driven automated evaluation data generation approach, and a systematic embodied capability taxonomy. 
A conceptual overview of Embodied Arena is shown in Figure~\ref{fig:Embodied Arena_overview}.

Embodied Arena evaluates both general large models and Embodied AI models, including leading commercial models and advanced academic models.
Embodied Arena is also eagerly calling for more open-source, closed-source models from multiple sources to join, with professional and user-friendly support by different means.
The evaluation data of Embodied Arena consists of (1) a diverse range of existing embodied benchmarks, which are carefully integrated and aligned by us, and (2) generative data powered by our LLM-driven automated generation pipeline.
Similarly, we also provide support for more benchmarks to join.
Hence, Embodied Arena keeps evolving the embodied evaluation data by integrating more benchmarks and generating new data.
With the evolving evaluation data, Embodied Arena conducts a comprehensive evaluation for each model based on the unified embodied evaluation system.
The evaluation results span three types of embodied tasks (i.e., Embodied Q\&A, Embodied Navigation, Embodied Task Planning), against the systematic embodied capability taxonomy (includes seven embodied capabilities with 25 fine-grained dimensions).
Finally, three types of leaderboards are summarized and presented for convenient and useful reference to both academia and industry.
Embodied Arena reacts in real-time to requests for evaluation and participation and updates the leaderboards and the evaluation system regularly.

Embodied Arena is designed with six core features that distinguish it as the comprehensive evaluation platform for Embodied AI models. These features address the fundamental challenges in Embodied AI evaluation while providing comprehensive support for the research community.
We highlight the six key features below:
\vspace{-0.2cm}
\begin{itemize}
    \item \textbf{Comprehensive Embodied Capability Taxonomy}: Embodied Arena introduces a systematic categorization spanning 7 core embodied capabilities decomposed into 25+ fine-grained dimensions, carefully refined from diverse embodied tasks and benchmarks to enable researchers to identify specific capability gaps and track progress across different aspects of Embodied AI.

    \item \textbf{Rich Model Support}: Our platform supports 30+ advanced models from 20+ leading research institutes worldwide, including general multimodal LLMs, specialized embodied models, and both open-source and commercial models through various access methods including API-based evaluation, parameter-based integration, and custom interfaces.

    \item \textbf{Modular Benchmark Integration}: Embodied Arena integrates 22+ evaluation benchmarks across three core domains with flexible extensibility through modular design that enables easy onboarding while maintaining consistent evaluation protocols as the platform evolves with field advancement.

    \item \textbf{Unified Evaluation Infrastructure}: The platform provides a standardized evaluation framework with uniform input/output formats, professional experiment management, and real-time leaderboard systems for transparent result presentation while ensuring consistent protocols and monthly updates.
    
    \item \textbf{High-quality Evaluation Datasets}: Embodied Arena maintains curated datasets continuously evolved through our LLM-driven automated generation pipeline, ensuring the scalability and diversity of embodied evaluation data while breaking manual construction bottlenecks.
    
    \item \textbf{Diverse Evaluation Methodologies}: Our platform employs complementary evaluation paradigms including accuracy-based QA assessment and interactive simulation-based testing, providing thorough assessment with flexibility across different benchmark characteristics for comprehensive embodied capability evaluation.
\end{itemize}

\section{Systematic Embodied Capability Taxonomy}
\label{sec:taxonomy}

\begin{figure}[t]
    \centering
    \includegraphics[width=1.0\linewidth]{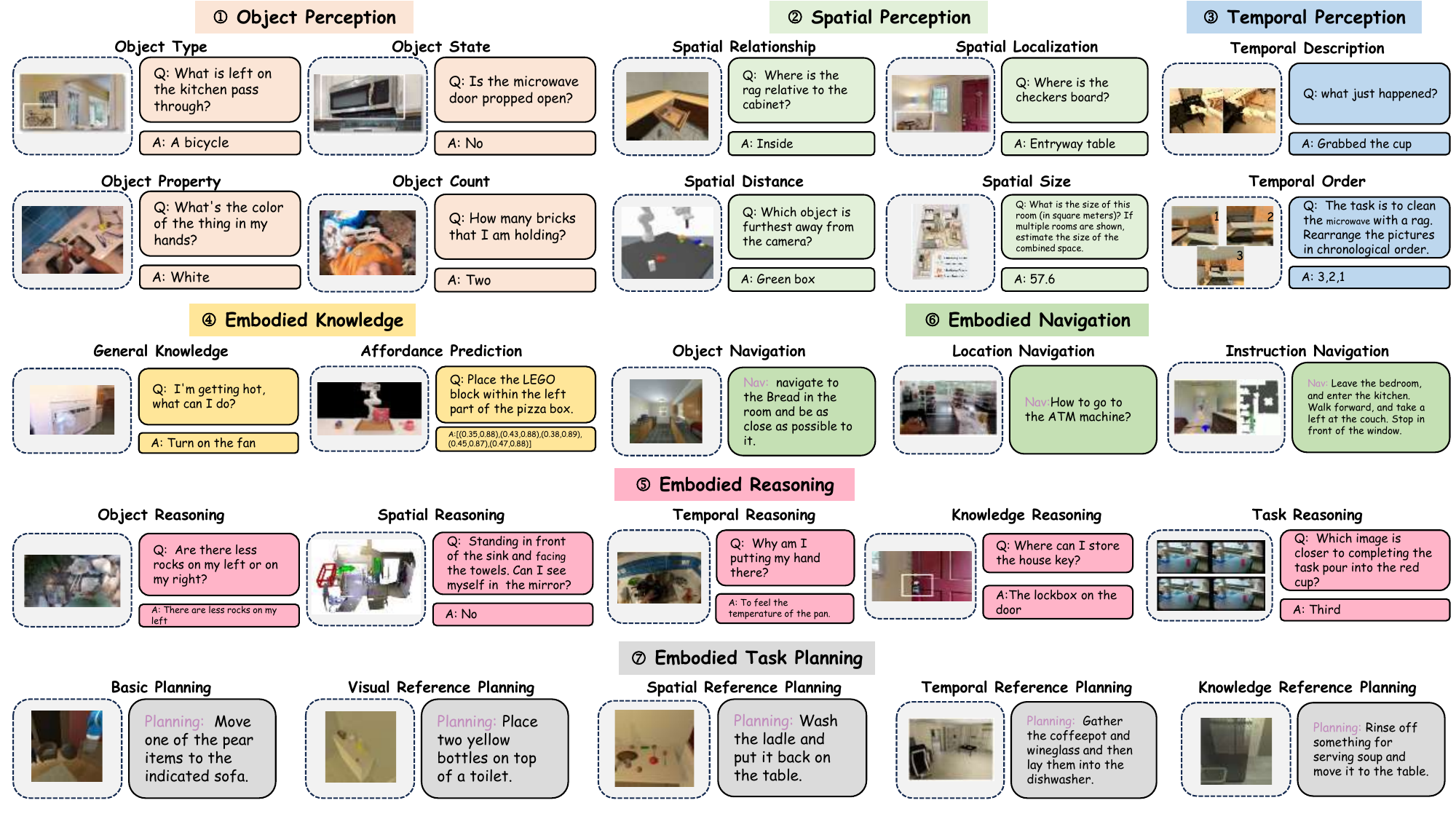}
    \caption{\textbf{The Systematic Embodied Capability Taxonomy and Exemplary Descriptions}. Embodied Arena encapsulates seven core capabilities: \textit{object perception}, \textit{spatial perception}, \textit{temporal perception}, \textit{embodied knowledge}, \textit{embodied reasoning}, \textit{embodied navigation}, and \textit{embodied task planning}, which contain 25 fine-grained capability dimensions in total.}
    \label{fig:core_cap}
\end{figure}

Embodied Arena uses a systematic taxonomy of capabilities potentially required for Embodied AI, drawing from cognitive psychology, human experience, and diverse existing tasks and benchmarks in the field~\citep{cheng2024egothinkevaluatingfirstpersonperspective,yang2024think}. Specifically, Embodied Arena considers seven core capabilities from low to high level: \textit{Object Perception}, \textit{Spatial Perception}, \textit{Temporal Perception}, \textit{Embodied Knowledge}, \textit{Embodied Reasoning}, \textit{Embodied Navigation}, and \textit{Embodied Task Planning}. 
More specifically, Object Perception, Spatial Perception, Temporal Perception, and Embodied Knowledge are viewed as the \textit{fundamental} embodied capabilities, responsible for multifaceted perception and low-level understanding.
Embodied Reasoning is treated as an \textit{advanced} embodied capability, as it is built upon the fundamental capabilities for understanding and conceiving solutions to complex questions and tasks.
It then moves on to \textit{downstream task-related} embodied capabilities, i.e., Embodied Navigation and Embodied Task Planning, at the high level of the taxonomy hierarchy.
Each core capability consists of multiple fine-grained capability dimensions. Figure~\ref{fig:core_cap} illustrates the 
typical task instance for each capability dimension.

\vspace{-0.6cm}
\paragraph{Object Perception} Recognizing objects via visual inputs is a fundamental capability for embodied models. Here we further divide the object perception into four fine-grained dimensions: \textit{Object Type}, recognizing specific categories of objects; \textit{Object Property}, determining physical properties of objects, e.g., color, shape, material, size, etc; \textit{Object State}, judging states of objects, e.g., open, closed, stationary, etc; \textit{Object Count}, recognizing the number of objects.

\vspace{-0.6cm}
\paragraph{Spatial Perception} Spatial perception capability is a vital core capability of embodied models. Accurate spatial perception is crucial for embodied agents to successfully perform tasks. Specifically, we further divide the spatial perception into four fine-grained dimensions: \textit{Spatial Relationship}, judging relative relationships (e.g., next to) of objects; \textit{Spatial Distance}, judging relative or absolute distances; \textit{Spatial Localization}, detecting the positions of objects; \textit{Spatial Size}, estimating the size of spaces, e.g., the size of rooms, etc.

\begin{table}[t]
\resizebox{\linewidth}{!}{
\begin{tabular}{|l|l|l|}
\hline
\textbf{Embodied Tasks} & \textbf{Embodied Benchmarks} & \textbf{Embodied Capabilities} \\ \hline
Embodied Q\&A & RoBoVQA~\citep{sermanet2024robovqa}& \begin{tabular}[c]{@{}l@{}}Temporal Perception: Temporal Description \\ Embodied Knowledge: Affordance Prediction \\ Embodied Reasoning: Task Reasoning\\ Embodied Task Planning: Basic Planning\end{tabular} \\ \hline
Embodied Q\&A & VSI-Bench~\citep{yang2024think} & \begin{tabular}[c]{@{}l@{}}Object Perception: Object Property, Object Count \\ Spatial Perception: Spatial Relationship, Spatial Distance, Spatial Size \\ Temporal Perception: Temporal Order \\ Embodied Task Planning: Basic Planning\end{tabular} \\ \hline
Embodied Q\&A & OpenEQA~\citep{OpenEQA2023} & \begin{tabular}[c]{@{}l@{}}Object Perception: Object Type, Object Property, Object State \\ Spatial Perception: Spatial Localization \\ Embodied Knowledge: General Knowledge\\ Embodied Reasoning: Spatial Reasoning, Knowledge Reasoning\end{tabular} \\ \hline
Embodied Q\&A & Where2Place~\citep{yuan2024robopoint} & Embodied Knowledge: Affordance Prediction \\ \hline
Embodied Q\&A & ERQA~\citep{geminiroboticsteam2025geminiroboticsbringingai} & \begin{tabular}[c]{@{}l@{}}Object Perception: Object Type, Object State \\ Spatial Perception: Spatial Relationship\\ Embodied Reasoning: Spatial Reasoning, Temporal Reasoning, Task Reasoning\end{tabular} \\ \hline
Embodied Q\&A & UniEQA~\citep{UniEQA2025} & \begin{tabular}[c]{@{}l@{}}Object Perception: Object Type, Object Property, Object State \\ Spatial Perception: Spatial Relationship\\ Temporal Perception: Temporal Description, Temporal Order,  \\ Embodied Knowledge: General Knowledge, Affordance Prediction \\ Embodied Reasoning: Object Reasoning, Task Reasoning \\ Embodied Navigation: Location Navigation\end{tabular} \\ \hline
Embodied Q\&A & VABench-Point~\citep{yuan2025seeingdoingbridgingreasoning} & Embodied Knowledge: Affordance Prediction \\ \hline
Embodied Q\&A & PhyBlock~\citep{ma2025phyblockprogressivebenchmarkphysical} & \begin{tabular}[c]{@{}l@{}}Object Perception: Object Type, Object Property, Object Count \\ Spatial Perception: Spatial Relationship, Spatial Distance \\ Temporal Perception: Temporal Order\\ Embodied Knowledge: Affordance Prediction\\ Embodied Reasoning: Spatial Reasoning, Knowledge Reasoning, Task Reasoning\\ Embodied Task Planning: Spatial Reference Planning\end{tabular} \\ \hline
Embodied Q\&A & MineAnyBuild~\citep{wei2025mineanybuild} & Embodied Reasoning: Spatial Reasoning, Knowledge Reasoning \\ \hline
Embodied Q\&A & ScanRefer~\citep{chen2020scanrefer} & Spatial Perception: Spatial Localization \\ \hline
Embodied Q\&A & Scan2Cap~\citep{chen2021scan2cap} & Spatial Perception: Spatial Relationship \\ \hline
Embodied Q\&A & ScanQA~\citep{azuma_2022_CVPR} & Spatial Perception: Spatial Localization \\ \hline
Embodied Q\&A & SQA3D~\citep{ma2022sqa3d} & Embodied Reasoning: Spatial Reasoning \\ \hline
Embodied Q\&A & Multi3DRefer~\citep{zhang2023multi3drefer} & Spatial Perception: Spatial Localization \\ \hline
Embodied Navigation & MP3D~\citep{Matterport3D} & Embodied Navigation: Object Navigation \\ \hline
Embodied Navigation & HM3D~\citep{ramakrishnan2021hm3d} & Embodied Navigation: Object Navigation \\ \hline
Embodied Navigation & EB-Navigation~\citep{yang2025embodiedbench} & Embodied Navigation: Object Navigation \\ \hline
Embodied Navigation & R2R-CE~\citep{yang2025embodiedbench} & Embodied Navigation: Instruction Navigation \\ \hline
Embodied Navigation & RxR-CE~\citep{yang2025embodiedbench,zhang2024navid} & Embodied Navigation: Instruction Navigation \\ \hline
Embodied Task Planning & ET-Plan-Bench~\citep{zhang2025etplanbenchembodiedtasklevelplanning} & Embodied Task Planning: Spatial Reference Planning, Temporal Reference Planning \\ \hline
Embodied Task Planning & EB-ALFRED~\citep{yang2025embodiedbench} & \begin{tabular}[c]{@{}l@{}}Embodied Task Planning: Basic Planning, Visual Reference Planning, \\ 
\quad \quad\quad \quad\quad \quad\quad \quad\quad \quad\quad \ Spatial Reference Planning, Knowledge Reference Planning\end{tabular} \\ \hline
Embodied Task Planning & EB-Habitat~\citep{yang2025embodiedbench} & \begin{tabular}[c]{@{}l@{}}Embodied Task Planning: Basic Planning, Visual Reference Planning, \\
\quad \quad\quad \quad\quad \quad\quad \quad\quad \quad\quad \  Spatial Reference Planning, Knowledge Reference Planning\end{tabular} \\ \hline
\end{tabular}}
\vspace{-0.1cm}
\caption{
\textbf{An overview of the mapping from existing embodied benchmarks to the three types of embodied tasks and the systematic embodied capability taxonomy in Embodied Arena.}}
\vspace{-0.2cm}
\label{tab:benchmark_mapping}
\end{table}

\vspace{-0.6cm}
\paragraph{Temporal Perception} Different from perceiving static semantics (such as object types and spatial relationships), temporal perception focuses on semantic content that changes over time. Here we investigate the temporal perception capability from two aspects, temporal description and temporal order. \textit{Temporal Description} recognizes the visual input contents related to the temporal dimension. \textit{Temporal order} judges the timestamp and sequential order of events based on visual inputs.

\vspace{-0.6cm}
\paragraph{Embodied Knowledge} Embodied knowledge refers to the basic cognitive capability of embodied models for the real world. Here we mainly focus on general knowledge and affordance prediction.
\textit{General knowledge} requires the embodied models to make judgments on general knowledge based on visual inputs. For example, refrigerators can keep food fresh. \textit{Affordance prediction} requires embodied models to infer possible object manipulations from visual inputs.

\vspace{-0.6cm}
\paragraph{Embodied Reasoning} Reasoning capability plays a crucial role in the decision-making process of complex embodied tasks. Based on basic perception and cognitive capabilities, we further divide the embodied reasoning into five fine-grained dimensions: \textit{Object Reasoning}, reasoning feasible actions on objects and comparing object properties based on object perception results; \textit{Spatial Reasoning}, reasoning about object accessibility, spatial inclusiveness, and spatial imagination, among other aspects, based on spatial perception results; 
\textit{Temporal Reasoning}, reasoning the causes and consequences of events based on temporal perception results; \textit{Knowledge Reasoning}, reasoning physical dynamics based on prior knowledge and visual inputs; \textit{Task Reasoning}, reasoning the type and location of task-related objects, task progress,  among other aspects, based on visual inputs and task instructions.

\vspace{-0.6cm}
\paragraph{Embodied Navigation} Navigation is a core embodied task in Embodied AI. Here we investigate the embodied navigation capability from three aspects: object navigation, location navigation, and instruction navigation.
\textit{Object Navigation} refers to the capability to navigate to a goal object from a start position. \textit{Location Navigation} refers to the capability to navigate to a goal location from a start position. \textit{Instruction Navigation} refers to the capability to follow a specified navigation instruction from a start position.

\vspace{-0.6cm}
\paragraph{Embodied Task Planning} Embodied Task Planning refers to decomposing the complex task into a reasonable sequence of sub-steps based on task instructions and visual inputs. Here we investigate the embodied planning capability from five aspects: basic planning, visual reference planning, spatial reference planning, temporal reference planning, and knowledge reference planning. \textit{Basic Planning} refers to the capability to decompose tasks where the instruction specifies object types. \textit{Visual Reference Planning} refers to the capability to decompose tasks where the instruction refers to objects using object properties, states, etc. \textit{Spatial Reference Planning} refers to the capability to decompose tasks with spatial constraints. \textit{Temporal Reference Planning} refers to the capability to decompose tasks with temporal constraints. \textit{Knowledge Reference Planning} refers to the capability to decompose tasks where the instruction refers to objects using object-related knowledge.

Based on our systematic taxonomy of embodied capabilities, we present a mapping in Table~\ref{tab:benchmark_mapping}, from existing embodied benchmarks to the three types of embodied tasks and the systematic embodied capability taxonomy in Embodied Arena.
We can observe that all the benchmarks focus on a single type of embodied tasks, and cover different parts of the capability dimensions in our systematic taxonomy.
This also indicates the unique value of Embodied Arena in providing a comprehensive evaluation for embodied models against complete embodied capability dimensions.

\section{Unified Embodied Evaluation System}
\label{sec:architecture}

In this section, we introduce the unified embodied evaluation system in Embodied Arena. This unified evaluation system aims to align the differences among existing embodied benchmarks and provide a standardized evaluation pipeline, thus closing the critical gap in cross-benchmark evaluation and comparison.
We detail the components of the system one by one in a logical order in the following.

\subsection{Tasks, Benchmarks, and Data}
In order to provide an in-depth and comprehensive evaluation of Embodied AI models, Embodied Arena currently covers three core types of embodied tasks: Embodied Question Answering, Embodied Navigation, and Embodied Task Planning. For each task, we carefully select high-quality evaluation benchmarks, which generally have broad academic influence and cover comprehensive and complementary capability dimensions.

Specifically, for Embodied Question Answering, we consider two types of benchmarks: 2D question answering and 3D question answering. The 2D question answering benchmarks include OpenEQA~\citep{OpenEQA2023}, VSI-Bench~\citep{yang2024think}, and ERQA~\citep{geminiroboticsteam2025geminiroboticsbringingai}, etc., and the 3D question answering benchmarks include the representative ScanQA~\citep{azuma_2022_CVPR}, Scan2Cap~\citep{chen2021scan2cap}, and SQA3D~\citep{ma2022sqa3d}, etc. For Embodied Navigation, we select the classic object navigation benchmarks MP3D~\citep{Matterport3D}, HM3D~\citep{ramakrishnan2021hm3d}, EB-Navigation~\citep{yang2025embodiedbench}, and the instruction navigation benchmarks R2R-CE~\citep{krantz_vlnce_2020} and RxR-CE~\citep{krantz_vlnce_2020,zhang2024navid}. For Embodied Task Planning, we consider EB-ALFRED, EB-Habitat~\citep{yang2025embodiedbench}, and ET-Plan-Bench~\citep{zhang2025etplanbenchembodiedtasklevelplanning}, which are more diverse in task types.

These benchmarks contain a total of more than 64k task instances. Among them, there are more than 48k embodied question-answer pairs, which are designed to comprehensively evaluate the model's performance in multiple embodied core capabilities. For embodied navigation tasks, Embodied Arena has accumulated more than 7k tasks, covering diverse navigation challenges of varying difficulty, aiming to comprehensively evaluate the embodied navigation capabilities of the model. In terms of embodied task planning, Embodied Arena provides more than 8k carefully designed tasks to examine the model's capability in the decomposition and execution of complex embodied tasks.
In addition, we provide a \textbf{Embodied Wiki} in the Embodied Arena platform, for convenient look-up and reference of the details of each benchmark.

 The current platform primarily focuses on perception, spatial reasoning, and high-level navigation and planning capabilities. While manipulation-related reasoning is included through QA and task planning Leaderboards, direct simulation-based manipulation tasks represent an important direction for future platform development. As the field evolves toward more sophisticated embodied agents, future extensions of the platform will incorporate more comprehensive manipulation tasks and closed-loop evaluation capabilities spanning the full perception-decision-action cycle.

Although Embodied Arena collects and integrates existing representative embodied benchmarks as mentioned above, the evaluation data in these benchmarks are static and finite.
To this end, Embodied Arena features a novel LLM-driven automated generation framework of embodied evaluation data. We defer the detailed introduction of it to Section~\ref{sec:generation}.

\subsection{Models}

Embodied Arena evaluates a comprehensive spectrum of AI models, ranging from general-purpose multimodal large language models to specialized Embodied AI models. Our platform encompasses both influential commercial models from leading technology companies and cutting-edge research models that represent the latest advances in Embodied AI. This diverse model ecosystem enables comprehensive cross-model comparison and provides valuable insights into the current landscape of embodied capabilities.

\subsubsection{General Multimodal Large Models}

General multimodal large language models represent foundation models with robust vision-language understanding capabilities, making them particularly well-suited for embodied question answering and high-level reasoning tasks. These models demonstrate exceptional language comprehension, sophisticated reasoning abilities, and excellent vision-language integration, delivering robust performance across diverse embodied scenarios. 

\vspace{-10pt}
\begin{itemize}
\item \textbf{OpenAI:} GPT-4o~\citep{openai2024gpt4o}, GPT-4o mini, o3, o4-mini~\citep{o3ando4mini}
\item \textbf{Google DeepMind:} Gemini-1.5-Pro, Gemini-1.5-flash~\citep{team2024gemini},  Gemini-2.5-Pro, Gemini-2.5-flash~\citep{gemini2024flash}
\item \textbf{Anthropic:} Claude-3.5-Sonnet, Claude-3.7-Sonnet~\citep{claude37sonnet2024}
\item \textbf{Alibaba Group:} Qwen-VL-Max, Qwen2-VL-7B-Instruct, Qwen2-VL-7B, Qwen2-VL-72B-Instruct~\citep{wang2024qwen2},  Qwen2.5-VL-3B-Instruct, Qwen2.5-VL-7B-Instruct, Qwen2.5-VL-7B, Qwen2.5-VL-72B-Instruct~\citep{bai2025qwen2}, mPlUG-Owl3~\citep{ye2024mplugowl3longimagesequenceunderstanding}
\item \textbf{ByteDance:} LongVA-7B~\citep{zhang2024long}, LLaVA-OneVision~\citep{li2024llava},  LLaVA-NeXT-Video~\citep{zhang2025llava}, pllava-7b~\citep{xu2024pllavaparameterfreellava}
\item \textbf{Meta AI:} Llama-3.2-11B-Vision-Instruct, Llama-3.2-90B-Vision-Instruct
\item \textbf{Shanghai AI Lab:} InternVL3~\citep{zhu2025internvl3},InterVL2.5~\citep{chen2024internvl}, InternVL2, 
\item \textbf{NVIDIA:} VILA-1.5~\citep{lin2024vila}
\item \textbf{Microsoft:} Phi-3-vision-128k-instruct~\citep{abdin2024phi3technicalreporthighly}
\item \textbf{ModelBest: } MiniCPM-V, MiniCPM-V 2.6~\citep{yao2024minicpm}
\end{itemize}
\vspace{-10pt}
These models excel in their strong linguistic understanding and reasoning capabilities, sophisticated vision-language integration, and particular suitability for complex question answering and high-level task reasoning scenarios. Their broad knowledge base and general-purpose design make them effective across multiple Embodied AI domains. However, while these general models provide strong fundamental capabilities, they often lack the specialized design and domain-specific optimizations required for complex embodied AI tasks, motivating the development of more targeted embodied AI models.

\subsubsection{Embodied AI Models}
Embodied AI models are specifically designed and optimized for embodied intelligence tasks, featuring enhanced spatial understanding, navigation capabilities, and physical interaction reasoning. Unlike general-purpose multimodal models, these models are comprehensively tailored for embodied scenarios through multiple dimensions: architecturally, they incorporate specialized components for spatial-temporal perception, affordance recognition, and action-oriented reasoning; in terms of training data, they leverage embodied-specific datasets including robotic trajectories, 3D scene interactions, and physical manipulation sequences; regarding training paradigms, they often employ supervised finetuning or reinforced post-training approaches adapted for embodied tasks. To better address the diverse requirements of embodied intelligence evaluation, these models are  categorized into 2D and 3D embodied models based on their primary application domains and the characteristics of their target environments.

\vspace{-0.5cm}
\paragraph{2D Embodied Models}
These models are specifically engineered for 2D visual reasoning benchmarks and embodied question answering tasks that operate within 2D visual representations~\citep{fu2024blink, chen2024spatialvlm}. They excel at processing egocentric viewpoints, understanding spatial relationships in 2D projections~\citep{liu2023visual, li2024topviewrs}, and reasoning about object interactions within constrained visual fields~\citep{cheng2024spatialrgpt, liao2024reasoning}. These models are primarily applied to VSI-Bench, ERQA, Where2Place, RoboVQA~\citep{sermanet2024robovqa}, and other 2D QA benchmarks~\citep{OpenEQA2023, azuma_2022_CVPR}, where they demonstrate superior performance in tasks requiring fine-grained spatial reasoning~\citep{cai2024spatialbot, ray2024sat}, temporal understanding from video sequences~\citep{bharadhwaj2024track2act, xu2024flow}, and affordance prediction in 2D visual contexts~\citep{yuan2024robopoint, nasiriany2024rt}.

\vspace{-10pt}
\begin{itemize}
\item \textbf{BAAI:} Navid~\citep{zhang2024navid}, UniNavid~\citep{zhang2024uni}, RoboBrain1.0-7B~\citep{ji2025robobrain}, RoboBrain2.0-7B, RoboBrain2.0-32B~\citep{baairobobrainteam2025robobrain20technicalreport}, MapNav~\citep{zhang2025quantum}
\item \textbf{Shanghai AI Lab:} VeBrain~\citep{luo2025visual}, VLN-R1~\citep{qi2025vln}, StreamVLN~\citep{wei2025streamvln}
\item \textbf{Tianjin University:} HuLE-Nav~\citep{han2024hule}, Embodied-R1~\cite{yuan2025embodiedr1reinforcedembodiedreasoning}
\item \textbf{University of Washington:} RoboPoint~\citep{yuan2024robopoint}
\item \textbf{Shanghai Jiao Tong University:} SpatialBot~\citep{cai2024spatialbot}
\item \textbf{The University of Hong Kong:} EmbodiedGPT~\citep{mu2023embodiedgpt}
\item \textbf{Google DeepMind:} LFG~\citep{shah2023navigation}, SpatialVLM~\citep{chen2024spatialvlm}
\item \textbf{Meta AI:} OVRL~\citep{yadav2022offlinevisualrepresentationlearning}, OVRL-v2~\citep{yadav2023ovrlv2simplestateofartbaseline}
\item \textbf{Peking University:} Space-R~\citep{ouyang2025spacer}, VoroNav~\citep{wu2024voronav}, InstructNav~\citep{long2024instructnav}
\item \textbf{Huawei Noah's Ark Lab:} Noah(UniE-VLM), OmniEVA
\item \textbf{NVIDIA:}  Cosmos-reason~\citep{liu2025cosmos}, NaVILA~\citep{cheng2024navila}
\item \textbf{Beihang University:} Robo-Refer~\citep{zhou2025roborefer}
\item \textbf{Boston University:} SAT~\citep{ray2024sat}
\item \textbf{University of California:} ESC~\citep{zhou2023escexplorationsoftcommonsense}
\item \textbf{University of California Berkeley:} VLMNav~\citep{goetting2024end}
\item \textbf{University of Groningen:} L3MVN~\citep{yu2023l3mvn}
\item \textbf{NYUAD Center for Artificial Intelligence and Robotics:} GAMap~\citep{yuan2024gamapzeroshotobjectgoal}
\item \textbf{Microsoft:} Magma~\citep{yang2025magma}

\end{itemize}

\vspace{-10pt}

\vspace{-0.4cm}
\paragraph{3D Embodied Models}
These advanced models are architecturally designed for comprehensive 3D scene understanding and complex spatial reasoning tasks that require full volumetric scene comprehension~\citep{hong20233d, chen2024grounded}. They incorporate sophisticated 3D feature extraction mechanisms, point cloud processing capabilities, and multi-view geometric reasoning to handle the inherent complexity of three-dimensional environments~\citep{zhu2024llava, li20243dmit}. These models excel at understanding object relationships in 3D space, reasoning about occlusions and spatial arrangements, and generating contextually aware descriptions of complex indoor scenes~\citep{yang20253d, qi2025gpt4scene}. They are primarily applied to ScanQA~\citep{azuma_2022_CVPR}, SQA3D~\citep{ma2022sqa3d}, Scan2Cap~\citep{chen2021scan2cap}, and other 3D QA benchmarks, where they demonstrate superior performance in tasks requiring dense captioning of 3D scenes, spatial localization within point clouds, and multi-hop reasoning across complex 3D spatial configurations~\citep{zheng2025video}.

\vspace{-10pt}
\begin{itemize}
\item \textbf{BIGAI:} LEO~\citep{huang2024embodiedgeneralistagent3d}
\item \textbf{Shanghai AI Lab:} Grounded 3D-LLM~\citep{chen2024grounded}, GPT4Scene~\citep{qi2025gpt4scene}
\item \textbf{Peking University}:  UniNavid~\citep{zhang2024uni}, Navid~\citep{zhang2024navid}
\item \textbf{The University of Hong Kong}: Video-3D LLM~\citep{zheng2025video}, LLaVA-3D~\citep{zhu2024llava}, GPT4Scene~\citep{qi2025gpt4scene}
\item \textbf{The Chinese University of Hong Kong}: Video-3D LLM~\citep{zheng2025video}
\item \textbf{UMass Amherst}: 3D-Mem~\citep{yang20253d}
\end{itemize}
\vspace{-10pt}

\subsection{Infrastructure}

Our unified evaluation infrastructure forms the backbone of Embodied Arena, ensuring consistent, reliable, and scalable assessment across all benchmarks through carefully designed system architecture and standardized protocols. The infrastructure is built with modularity, extensibility, and reproducibility as core design principles.

\vspace{-0.5cm}
\paragraph{Standardized Evaluation Framework} The platform implements a standardized evaluation framework with uniform input/output formats that enable seamless comparison across diverse models and benchmarks. This framework abstracts away benchmark-specific implementation details while preserving the unique characteristics of each evaluation task. The standardized interface supports various model access methods, including API-based evaluation for commercial models, parameter-based integration for open-source models, and custom interfaces for specialized architectures.

\vspace{-0.5cm}
\paragraph{Flexible Model Integration} Embodied Arena supports comprehensive evaluation of models from different sources (open-source, commercial) through various access methods (model parameters, API endpoints, custom implementations). This flexibility ensures broad accessibility and participation while maintaining evaluation consistency and fairness across different model types and deployment scenarios.

\vspace{-0.5cm}
\paragraph{Professional Management:}
The infrastructure includes comprehensive experiment tracking and management capabilities that provide detailed performance analysis and ensure reproducible evaluation results. Each evaluation run is meticulously logged with complete metadata including model configurations, benchmark parameters, execution environment details, and performance metrics.

\subsection{Evaluation Methods}

Embodied Arena extensively supports the comprehensive evaluation of models from different sources (open-source, commercial) by different means (model parameters, API), offering flexibility and convenience for users to join. The platform leverages a diverse range of well-curated Embodied AI benchmarks, ensuring high alignment with canonical evaluation methods and the best completeness compared to prior works.

\subsubsection{Evaluation Metrics}
During the evaluation phase, we select the corresponding evaluation metric based on the characteristics of the benchmark itself, which generally include the following types:

\textbf{Embodied Question Answering:}
\vspace{-10pt}
\begin{itemize}
\item \textbf{Exact Matching Accuracy}: Applied to benchmarks requiring precise categorical responses such as VSI-Bench, Where2Place, and ERQA~\citep{du2024embspatial}. This metric evaluates the model's ability to provide accurate factual answers and correct spatial reasoning outputs.
\item \textbf{Fuzzy Matching Accuracy}: Employed for benchmarks involving natural language generation and open-ended responses:
    \begin{itemize}
    \item \textit{Rule-based Metrics} (CIDEr, BLEU, ROUGE, MRA): Applied to benchmarks like RoboVQA~\citep{sermanet2024robovqa}, Scan2Cap~\citep{chen2021scan2cap}, and ScanQA~\citep{azuma_2022_CVPR} for evaluating generated descriptions and spatial explanations
    \item \textit{LLM-based Evaluation}: Utilized for benchmarks such as OpenEQA~\citep{OpenEQA2023} and UniEQA~\citep{UniEQA2025}, leveraging large language models to assess semantic correctness and reasoning quality in generated responses~\citep{zheng2023judging}
    \end{itemize}
\end{itemize}

\vspace{-5pt}
\textbf{Embodied Navigation Evaluation:}
\vspace{-10pt}
\begin{itemize}
\item \textbf{Success Rate}: Primary metric for navigation benchmarks including EB-Navigation, R2R-CE~\citep{krantz_vlnce_2020}, and RxR-CE, measuring the percentage of successfully completed navigation episodes
\item \textbf{Path Length Weighted Success Rate (SPL)}: Evaluates navigation efficiency by considering both success and path optimality
\end{itemize}

\vspace{-5pt}
\textbf{Embodied Task Planning Evaluation:}
\vspace{-10pt}
\begin{itemize}
\item \textbf{Task Completion Success Rate}: Applied to benchmarks such as EB-ALFRED, EB-Habitat, and ET-Plan-Bench~\citep{zhang2025etplanbenchembodiedtasklevelplanning}, measuring the percentage of successfully completed task sequences
\end{itemize}

\vspace{-5pt}

\subsubsection{Scoring Rules}
The scoring rules for the embodied capability leaderboards and the embodied task leaderboards are as follows:

\textbf{Embodied Task Leaderboards:} Given $N$ benchmarks, let there be a benchmark $B^n$ $(n=1,2,\cdots, N)$ consisting of $M$ fine-grained original capability dimensions. For each capability dimension $m$ $(m=1,2,\cdots,M)$, $k^n_m$ denotes the total number of questions in the $m$-th capability dimension, and $c^n_m$ is the number of questions answered correctly in the $m$-th capability dimension. Each question has a score within $[0,1]$.

\vspace{-10pt}
\begin{itemize}
\item \textbf{Score Calculation for a Single Benchmark}
    \begin{itemize}
    \item \textit{Capability Dimension Score} $S^n_m$: $S^n_m = \frac{c^n_m}{k^n_m}\times 100$, where $c^n_m\in [0,k^n_m]$ and $S^n_m\in [0,100]$.
    \item \textit{Total Score of All Capability Dimensions} $A^n_{total}$: $A^n_{total} = \frac{1}{M}\sum_{m=1}^{M}S^n_m$.
    \end{itemize}
\item \textbf{Total Score across $N$ Benchmarks}: $B_{total} = \frac{1}{N}\sum_{n=1}^{N}A^n_{total}$.
\end{itemize}

\vspace{-10pt}
\textbf{Embodied Capability Leaderboards:} Given $N$ benchmarks, let there be a benchmark $B^n$ $(n=1,2,\cdots, N)$ with $M^n$ original capability dimensions. Our taxonomy defines $D=25$ fine-grained capability dimensions and $P=7$ core capabilities. Let $\phi: (n,i) \rightarrow j$ be the mapping function from the $i$-th original dimension of benchmark $B^n$ to the $j$-th taxonomy dimension. For each taxonomy dimension $j$ $(j=1,2,\cdots,D)$, let $k^n_j$ and $c^n_j$ denote the total number of questions and correctly answered questions respectively, aggregated from all original dimensions of benchmark $B^n$ that map to dimension $j$.

\vspace{-10pt}
\begin{itemize}
\item \textbf{Score Calculation for a Single Benchmark}
    \begin{itemize}
    \item \textit{Fine-grained Capability Dimension Score} $S^n_j$: $S^n_j = \frac{c^n_j}{k^n_j}\times 100$ if $k^n_j > 0$, otherwise undefined.
    \item \textit{Core Capability Dimension Score} $C^n_p$: $C^n_p = \frac{1}{|I_p|}\sum_{j \in I_p} S^n_j$, where $I_p$ is the set of fine-grained dimensions belonging to core capability $p$.
    \item \textit{Total Score of All Capability Dimensions} $A^n_{total}$: $A^n_{total} = \frac{1}{|J^n|}\sum_{j \in J^n} S^n_j$, where $J^n$ is the set of taxonomy dimensions covered by benchmark $B^n$.
    \end{itemize}
\item \textbf{Total Score of $N$ Benchmarks on Fine-grained Capability Dimension $j$}: $T^j_{total} = \frac{\sum_{n=1}^{N} c^n_j}{\sum_{n=1}^{N} k^n_j}\times 100$ (only for benchmarks where $k^n_j > 0$).
\item \textbf{Total Score of $N$ Benchmarks for All Capability Dimensions} $B_{total}$: $B_{total} = \frac{1}{D}\sum_{j=1}^{D} T^j_{total}$ (only including dimensions with valid scores).
\end{itemize}

\subsection{Leaderboards}

Embodied Arena features a comprehensive leaderboard system designed to provide clear, actionable insights into model performance across different perspectives and granularities.
Through comprehensive evaluation among the growing model population based on evolving evaluation data, Embodied Arena publishes three types of leaderboards, i.e., Embodied Q\&A, Embodied Navigation, Embodied Task Planning, with two orthogonal views, i.e., \textit{benchmark view} and \textit{capability view}.
The benchmark view presents the ranking of models on each benchmark, which is convenient for academic researchers to quote and compare with in their research works; while the capability view instead presents the ranking of models against each embodied capability in the systematic taxonomy, providing an up-to-date overview of embodied capabilities of advanced models. 
Moreover, to ensure evaluation integrity and community engagement, our leaderboard system implements structured monthly updates with transparent submission policies and real-time performance tracking.
The ultimate aim of Embodied Arena is to facilitate the establishment of clear research veins and help to identify critical research problems, thereby propelling research progress in the field of Embodied AI.

\vspace{-0.5cm}
\paragraph{Three Types of Leaderboards}  Embodied Arena establishes comprehensive evaluation through three specialized leaderboards targeting distinct Embodied AI domains. The \textit{Embodied Question Answering} leaderboard evaluates models across visual reasoning tasks, assessing capabilities from 2D visual understanding to 3D spatial comprehension. The \textit{Embodied Navigation} leaderboard focuses on spatial movement and pathfinding abilities, evaluating object navigation, location navigation, and instruction-following capabilities. The \textit{Embodied Task Planning} leaderboard assesses high-level reasoning and decomposition skills, examining models' abilities to break down complex tasks into executable sequences. Together, these leaderboards provide comprehensive coverage of core Embodied AI competencies.

\vspace{-0.5cm}
\paragraph{Two Orthogonal Comparison Views} Embodied Arena provides complementary insights through its dual-view leaderboard system. The \textit{benchmark view} presents model rankings on individual benchmarks, enabling direct comparison and facilitating academic citation. In contrast, the \textit{capability view} aggregates performance across our seven core embodied capabilities (detailed in \cref{sec:taxonomy}), providing strategic insights into model strengths and weaknesses. Together, these orthogonal views deliver both granular benchmark-specific analysis and holistic capability assessment, enabling targeted improvement while maintaining systematic understanding of Embodied AI advancement.

\vspace{-0.5cm}
\paragraph{Update and Submission Policies} Embodied Arena maintains evaluation integrity through structured update protocols and submission policies. The leaderboards are updated monthly with performance snapshots taken on the first working day of each month, ensuring consistent and timely community engagement. To ensure fairness, each organization is limited to one evaluation submission per month, with results typically processed and updated within seven working days of submission. This systematic approach provides dynamic performance tracking while maintaining evaluation reliability and preventing potential gaming of the leaderboard system through excessive submissions.

\vspace{-0.5cm}
\paragraph{Platform Accessibility and Community Engagement} Embodied Arena operates as an open evaluation platform designed to foster community-driven advancement in embodied AI research. The platform provides multiple pathways for researcher participation: \textit{open evaluation access} allows researchers to submit models for assessment through standardized API interfaces, \textit{benchmark contribution} enables community members to propose and integrate new evaluation tasks following our established guidelines, and \textit{transparent methodology} ensures all evaluation protocols and baseline implementations remain publicly accessible. Through comprehensive documentation, integration templates, and testing scripts, Embodied Arena lowers barriers to participation while maintaining evaluation consistency. This open architecture not only democratizes access to comprehensive embodied AI evaluation but also enables the platform to evolve continuously with field advancement through community contributions and feedback.

\section{Automated Data Generation for Embodied AI Evaluation}
\label{sec:generation}

Current evaluation benchmarks for embodied tasks suffer from fundamental limitations in adaptability, scalability, and task diversity, which restrict their effectiveness. In contrast, \textbf{Embodied Arena} is designed as a continuously evolving evaluation platform—one that actively identifies model weaknesses and autonomously generates new, targeted data to maintain the comprehensiveness and cutting-edge nature of the benchmark over time.
Specifically, we identify three core limitations in existing benchmarks:

\begin{figure}[t]
    \centering
    \includegraphics[width=0.75\linewidth]{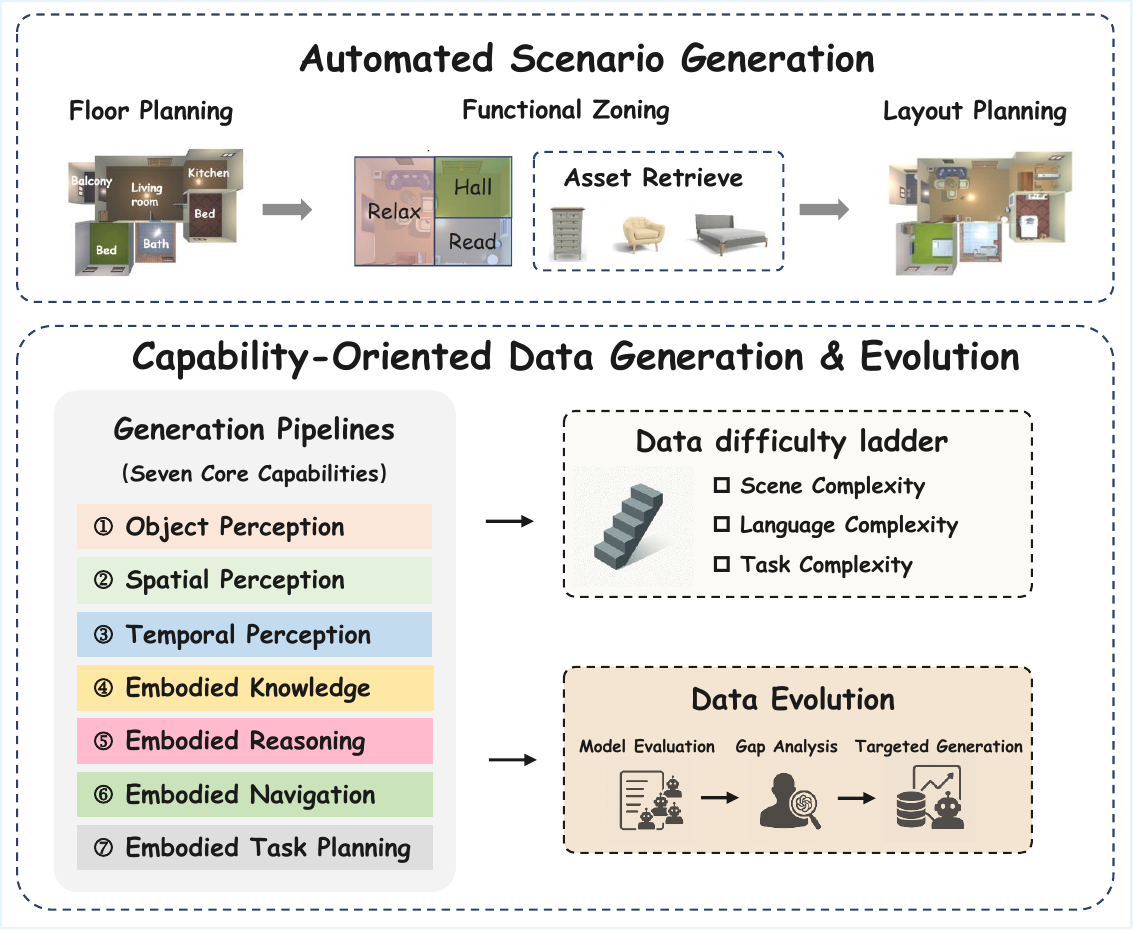}
    \vspace{-0.3cm}
    \caption{\textbf{Illustration of the automated data generation pipeline.} The pipeline includes two modules: Automated Scenario Generation and Capability-Oriented Data Generation \& Evolution. The former is responsible for generating diverse and realistic high-fidelity scenarios, while the latter builds generation pipelines to ensure the continuous evolution of the evaluation set.}
\label{fig:data gen}
\end{figure}

\vspace{-0.5cm}
\begin{itemize}
  \item \textbf{Static evaluation}: Conventional benchmarks are typically constructed once and remain fixed, without adapting to model performance. This static nature introduces the overfitting risk, whereby agents achieve high performance on existing data but fail to generalize to out-of-distribution (OOD) data.
  \item \textbf{Limited scalability}: Current benchmarks rely heavily on manual annotation, which is both labor-intensive and time-consuming, rendering it infeasible to collect large-scale evaluation datasets efficiently.
  \item \textbf{Limited diversity}: Most handcrafted data focus on a small set of tasks, making it difficult to evaluate the broad spectrum of embodied capabilities or generalization to novel tasks.
\end{itemize}
\vspace{-0.3cm}

To overcome these challenges, we introduce an LLM-driven evaluation data generation framework built on high-fidelity simulation. Our framework consists of two key modules: \textit{Automated Scenario Generation}, which constructs realistic and diverse simulation environments, and \textit{Capability-Oriented Data Generation \& Evolution}, which establishes data generation pipelines and continuously injects targeted data to adapt the evaluation set based on the model limitations.

\subsection{Automated Scenario Generation}
Since embodied agents operate in complex physical environments, the ability to simulate realistic and diverse scenes is essential for evaluating their embodied capabilities across a broad range of tasks.
To address this, the Automated Scenario Generation Module is designed to automatically build multi-room indoor environments through a structured, hierarchical process that mirrors real-world scene building process. 
The generation pipeline consists of three stages: (1) \textit{Floor Planning}, which defines room types and their spatial relationships to ensure logical connectivity and functional plausibility; (2) \textit{Functional Zoning}, which divides each room into activity-specific zones (e.g., cooking, dining, storage); and (3) \textit{Layout Planning}, which populates each zone with diverse assets and applies layout optimizations.

To ensure the generated scenes are aligned with human common sense and real-world affordances, we leverage large language models (LLMs) and vision–language models (VLMs) throughout the generation process. These models guide decisions such as object spatial relations and scene semantics, enabling the creation of environments that are not only diverse but also semantically coherent.
Once the scene is constructed, a high-fidelity rendering pipeline produces rich outputs, including RGB images, depth maps, and a structured object graph. Domain randomization techniques are applied to introduce variability in textures, lighting, and viewpoints, enhancing generalization for downstream tasks. Furthermore, the module offers a \textit{targeted scene} mode, in which users provide high-level descriptors --- such as “cluttered kitchen with partially hidden utensils” or “open-concept living room featuring scattered numeric signs”. The system then samples room layouts, places assets, and applies refinements to construct these concrete indoor environments, yielding reproducible scenes that match the specified requirements.

\subsection{Capability-Oriented Data Generation \& Evolution}
To generate datasets for the seven core embodied capabilities, we design simulator-driven procedural pipelines.
Each pipeline (i) defines a capability-specific task template and specification, (ii) loads scenes and task-relevant assets, (iii) executes scripted procedures, (iv) leverages the simulator’s privileged access to automatically extract object types, positions, attributes, and other ground-truth annotations as the basis for dataset construction, and (v) performs automated filtering and selection to retain only unambiguous, high-quality data before storage.

Building on these pipelines, we introduce the difficulty ladder that generates data along three dimensions:

\vspace{-0.3cm}
\begin{itemize}
    \item Scene Complexity (number of objects, degree of occlusion)
    \item Language Complexity (instruction length, semantic complexity)
    \item Task Complexity (Horizon length, temporal dependencies)
\end{itemize}
\vspace{-0.3cm}

At each level, we generate visual-instruction-answer triplets. During evaluation, agents can unlock levels sequentially, enabling granular diagnosis of strengths and weaknesses and providing a built-in curriculum for progressive fine-tuning.

Procedural data generation relies on privileged information obtained from simulation, which ensures correctness but may also introduce ambiguous cases that lead to model hallucinations. For example, certain assets in corners may be recognizable in simulation even when only a small portion is exposed, yet such cases remain challenging for human observers.
To ensure data quality, we adapt a sampling-based inspection method, where human evaluation is used to filter the data and remove cases that are difficult to discern for the human eye. Although this process introduces a certain degree of manual overhead, it offers a more reliable safeguard for data accuracy and validity.

To support the long-term effectiveness of Embodied Arena, we introduce a data evolution mechanism driven by model performance analysis. By regularly analyzing model capability, we generate the related data via the aforementioned pipelines.
These targeted additions enrich the evaluation set with fresh, challenging data tailored to current model limitations.
In this way, Embodied Arena evolves alongside model progress, maintaining comprehensiveness while continuously advancing in difficulty and evaluation value.

\vspace{-2mm}
\section{Insights from Embodied Arena Evaluation and Leaderboards}
\label{sec:insights}

Through comprehensive evaluation across the diverse benchmarks and model ecosystem on our platform, Embodied Arena reveals several major insights that illuminate the current state and future directions of Embodied AI research. These findings emerge from a systematic analysis of performance patterns across 30+ models and 22+ benchmarks, providing empirical evidence for understanding the fundamental capabilities and limitations of contemporary embodied intelligence systems.

\begin{figure}[t]
    \centering
    \includegraphics[width=0.98\linewidth]{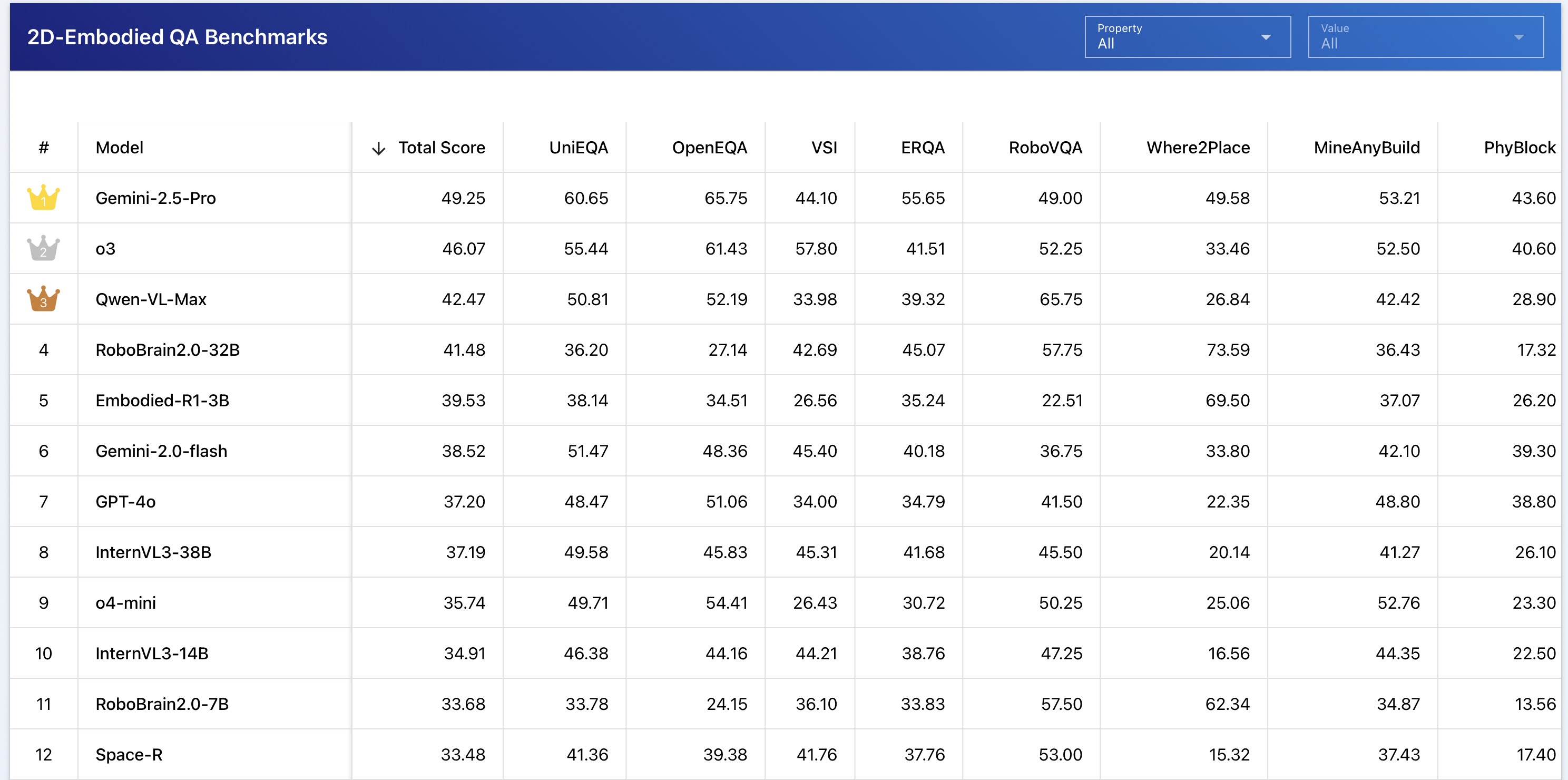}
    \caption{\textbf{A Screenshot of 2D Embodied QA Leaderboard from Embodied Arena.} The leaderboard shows performance rankings across multiple 2D embodied question answering benchmarks, with large-scale general multimodal models generally achieving higher overall scores than specialized embodied models. This convenient web interface enables researchers to easily analyze model performance patterns and compare capabilities across different approaches.}
    \label{fig:finding1_2d_qa}
    \vspace{-5pt}
\end{figure}

\begin{itemize}[leftmargin=0pt]

\item \textbf{Finding 1:} \textbf{Embodied models surpass general models of similar sizes on specialized benchmarks, while top-tier closed-source general models achieve strong overall performance with large model size and massive training data.}

\vspace{0.1cm}
\textbf{Core Finding:} As illustrated in Figure~\ref{fig:finding1_2d_qa}, our evaluation reveals a significant phenomenon that massive general models achieve strong performance through large model size and broad knowledge, while specialized models excel through targeted embodied training. Large commercial general models, e.g., GPT-o3~\citep{openai2025o3}, Gemini-2.5-Pro~\citep{gemini2024flash}, Claude-3.7~\citep{claude37sonnet2024}, leverage their massive scale to achieve overall 10-20\% performance advantages across most benchmarks, demonstrating that model size and extensive pre-training data clearly matter. However, when we compare models of similar scales, specialized embodied models consistently outperform the general multimodal models. RoboBrain2.0-32B~\citep{RoboBrain2.0TechnicalReport} achieves 73.59\% on Where2Place versus GPT-o3's 33.46\%, while navigation specialists like StreamVLN~\citep{wei2025streamvln} and UniNavid~\citep{zhang2024uni} reach 30-57\% success rates while the general models of similar sizes such as InternVL3~\citep{zhu2025internvl3} achieve less than 10\% success rates.

\textbf{In-depth Analysis:} 
\emph{Massive model scale powered by large-scale pre-training results in clear advantages} --- commercial models leverage massive parameter scales (likely hundreds of billions to trillions) trained on internet-scale datasets, providing both reasoning capacity and extensive world knowledge. However, specialized embodied data also demonstrates remarkable power. Models like RoboBrain2.0~\citep{RoboBrain2.0TechnicalReport}, Embodied-R1, and RoboPoint~\citep{yuan2024robopoint} show that \emph{fine-tuning small-scale open-source general models with high-quality embodied datasets can produce dramatic performance gains}. By incorporating specialized training data focused on spatial identification, affordance prediction, and manipulation sequences, these models vastly outperform their original versions and can even match large commercial models on specific embodied benchmarks. This effect is particularly evident in navigation models like StreamVLN~\citep{wei2025streamvln} and NaVILA~\citep{zheng2024towards}, which use vision-language-action training paradigms to develop capabilities that general pre-training cannot provide. This demonstrates that \emph{targeted incorporation of embodied-specific data represents a viable pathway to achieving competitive performance even with constrained computational resources.} 
This \emph{post-training optimization pathway} promises to be a key research direction for enabling smaller specialized embodied models to match or surpass large closed-source general models through better data quality and targeted architectural innovations.

\begin{figure}[t]
    \centering
    \includegraphics[width=0.98\linewidth]{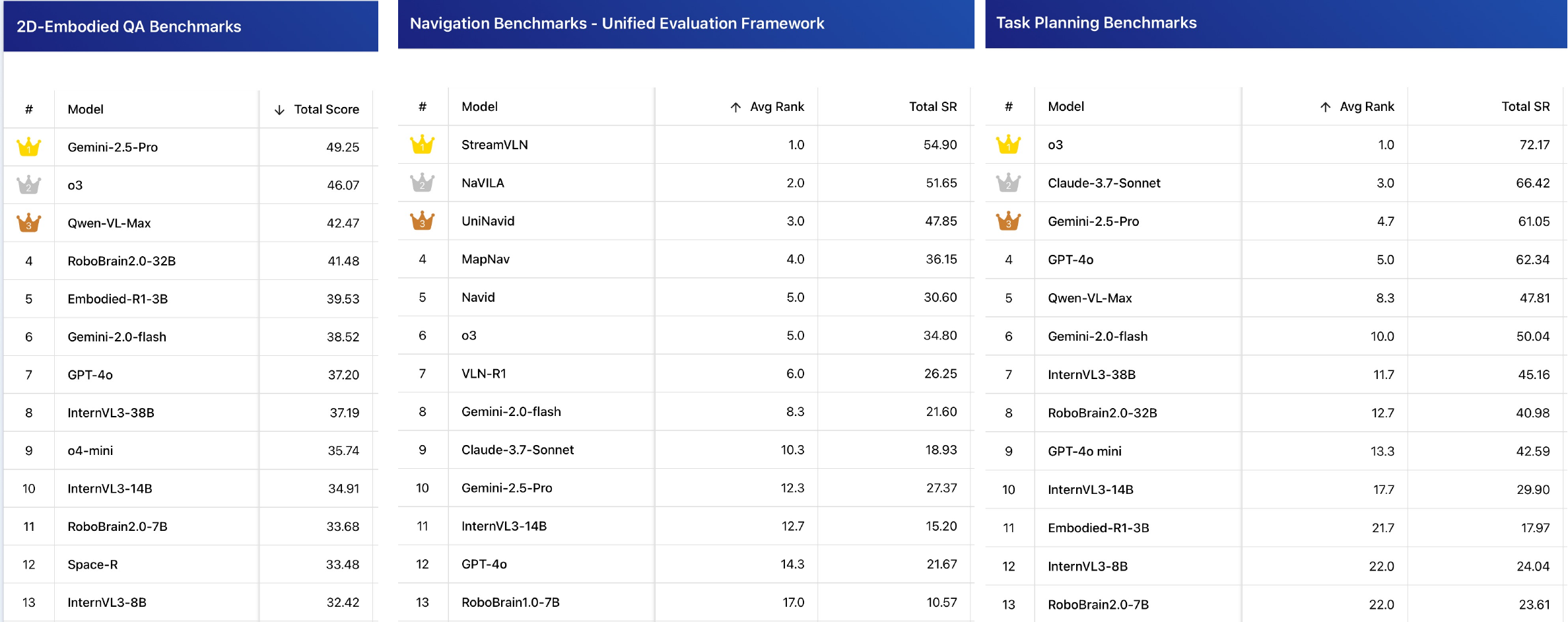}
    \caption{\textbf{A Screenshot of Cross-Benchmark Performance Variation from Embodied Arena.} The figure displays ranking comparisons across 2D Embodied QA, Navigation, and Task Planning leaderboards, revealing significant performance fluctuations where models excel in specific domains but struggle to maintain consistent rankings across all task types. This comprehensive view from our web platform facilitates easy analysis of model strengths and weaknesses across different embodied capabilities.}
    \label{fig:finding2_cross_benchmark}
    \vspace{-5pt}
\end{figure}

\vspace{0.2cm}
\item \textbf{Finding 2:} \textbf{Individual benchmarks with limited scope are insufficient and biased for embodied evaluation. Embodied models exhibit more or less overfitting on benchmark-specific data rather than developing comprehensive embodied capabilities.}

\vspace{0.1cm}
\textbf{Core Finding:} Models exhibit dramatic performance variations across different benchmarks, revealing fundamental limitations in both current evaluation benchmarks in Figure~\ref{fig:finding2_cross_benchmark}. For instance, RoboBrain-v1-7B~\citep{ji2025robobrain} achieves top performance on RoboVQA~\citep{sermanet2024robovqa} across all metrics but performs poorly on spatial understanding benchmarks like Where2Place~\citep{yuan2024robopoint} and VSI-Bench~\citep{yang2024think}. Similarly, while specialized navigation models dominate VLN tasks with success rates above 50\%, they struggle with basic question answering tasks. Only a few large-scale models like Gemini-2.5-Pro~\citep{gemini2024flash}, GPT-o3~\citep{openai2024gpt4o}, and InternVL3-38B~\citep{zhu2025internvl3} maintain relatively balanced performance across question answering, navigation, and task planning. 

\vspace{0.1cm}
\textbf{In-depth Analysis:} This performance inconsistency across benchmarks directly validates the core motivation for establishing our comprehensive embodied leaderboard system, which aims to provide holistic model evaluation beyond isolated task performance. \emph{Each benchmark evaluates only limited capability dimensions, and no single benchmark currently covers all embodied capabilities comprehensively}, causing significant ranking fluctuations as models demonstrate varying strengths across embodied AI capabilities --- spatial reasoning benchmarks like Where2Place~\citep{yuan2024robopoint} favor affordance prediction training while task planning benchmarks like EB-ALFRED~\citep{yang2025embodiedbench} advantage instruction following abilities. 
Moreover, \emph{current embodied models exhibit concerning overfitting phenomena where capabilities appear artificially enhanced through injecting benchmark-correlated data for specialized ability improvements} --- performance on one benchmark can be improved simply by adding specific related training datasets, but this comes at the expense of performance on other benchmarks rather than achieving true comprehensive enhancement of embodied capacity.
To this end, one possible solution is developing \emph{automated data generation systems} that create diverse scenarios and tasks for comprehensive evaluation and training. These systems will enable \emph{unified evaluation paradigms} that assess models' ability to integrate perception, temporal reasoning, and planning in realistic scenarios, providing robust assessment that resists benchmark-specific overfitting.

\vspace{0.2cm}
\item \textbf{Finding 3:}
\textbf{The embodied reasoning capabilities of models are strongly dependent on their fundamental embodied capabilities. Among the five fundamental embodied capabilities, object perception and spatial perception turn out to be the major bottlenecks.}

\vspace{0.1cm}
\textbf{Core Finding:}
The comprehensive evaluation results across multiple benchmarks indicate that the defects in the models' fundamental embodied capabilities directly restrict their performance in the advanced reasoning capability. Specifically, the models' fundamental embodied capabilities (object perception, spatial perception, temporal perception, and embodied knowledge) show a significant positive correlation with advanced reasoning capability (Spearman's rank correlation coefficient $\rho=0.80$, p<0.0001), and each fundamental embodied capability also exhibits a significant positive correlation with advanced reasoning capability ($\rho$ ranging from 0.68 to 0.77, p<0.0001). Meanwhile, the models' performance on advanced reasoning capability (with an average score of 33.64) is generally worse than their overall performance on fundamental embodied capabilities (with an average score of 38.84). Among the fundamental embodied capabilities, the models' object perception (average score 38.33) and spatial perception (average score 28.62) capabilities are the worst.
These results collectively reveal the deep dependence of the model's advanced reasoning abilities on its fundamental embodied capabilities.

\vspace{0.1cm}
\textbf{In-depth Analysis:} \emph{The embodied capabilities of the evaluated models exhibit a hierarchical decline}, primarily stemming from three core factors: 
first, there is \emph{a structural imbalance in the pre-training data} for different embodied capabilities, with labeled data related to reasoning capabilities being particularly scarce; second, \emph{the deficiency of specific embodied capabilities easily leads to performance declines of models in other ones} due to the dependencies among embodied capabilities; third, \emph{the reasoning capabilities of embodied models are still lacking} while the methods for enhancing embodied reasoning capabilities have not been well studied yet.

\vspace{0.2cm}
\item \textbf{Finding 4:}
\textbf{The fundamental and advanced embodied capabilities of models are significantly positively correlated with their performance on downstream embodied tasks.
Furthermore, in an end-to-end manner, there is a moderate correlation between the model's embodied capabilities and downstream task performance. In contrast, in a task-oriented agentic framework manner, there is a strong correlation between the model's embodied capabilities and downstream task performance.}

\vspace{0.1cm}
\textbf{Core Finding:} 
Based on the comprehensive ranking of the models in embodied capabilities (e.g., object perception, spatial perception, temporal perception, embodied knowledge, and embodied reasoning) and downstream tasks (e.g., embodied navigation, embodied task planning), we find that models' embodied capabilities are highly positively correlated with downstream task performance (Spearman's rank correlation coefficient $\rho=0.80$, $p<0.0001$).
Moreover, each embodied capability shows a significant positive correlation with downstream task performance, with correlation coefficients $\rho$ ranging from 0.73 to 0.83.
Among these, embodied knowledge demonstrates the strongest positive correlation ($\rho=0.83$, $p<0.0001$), and the remaining capabilities show stable correlation coefficients around 0.75. All p-values for the above analyses are less than 0.0001, reaching an extremely high level of statistical significance. 
Notably, the strength of this correlation is significantly influenced by how the model is applied in downstream tasks: in an end-to-end manner, the model's embodied capabilities and downstream task performance exhibit only a moderate correlation ($\rho=0.40$, $p>0.08$), which does not reach statistical significance; However, in a task-oriented agentic framework manner (i.e., integrating the general models into agentic frameworks targeted for downstream embodied tasks), the two exhibit a strong positive correlation ($\rho=0.79$, $p<0.0001$). In addition, we find that when general models are applied in an end-to-end manner, their overall success rate in navigation tasks is only 5.80\%. In contrast, when using task-oriented agentic frameworks (such as EmbodiedBench~\citep{yang2025embodiedbench} and ET-Plan-Bench~\citep{zhang2025etplanbenchembodiedtasklevelplanning}), general models' success rates in navigation and task planning tasks increase to 36.21\% and 40.08\%, respectively.

\vspace{0.1cm}
\textbf{In-depth Analysis:} These evaluation results reveal two key insights about how to convert the embodied capabilities of models into performance on downstream embodied tasks.
On one hand, the embodied capabilities of models are the foundation for the performance in downstream embodied tasks. Therefore, further enhancing the embodied capabilities like perception and reasoning is an essential necessity.
On the other hand, compared with the end-to-end approach, \emph{task-oriented agentic framework provides an effective pathway} for better utilizing the fundamental embodied capabilities of general models in downstream embodied tasks, although these \emph{agentic frameworks usually require manual design and lack generality}. 
Leveraging learning-based methods like RL to optimize or generate the agentic framework can be promising to facilitate the transformation of embodied capabilities into task performance.

\vspace{0.2cm}
\item \textbf{Finding 5:} 
\textbf{The scaling law for embodied tasks has yet to emerge. Larger model size does not consistently lead to stronger embodied capabilities, although a positive correlation exists locally for specific models and capabilities. More embodied data leads to improved task-specific performance, albeit with increased overfitting.}

\vspace{0.1cm}
\textbf{Core Finding:} 
For the same model architecture, increasing the number of parameters can improve performance on specific embodied benchmarks. General multimodal models show more consistent evidence of this, 
whereas for embodied models, this phenomenon is inconsistent across models and capabilities.
For instance, InternVL3-38B > InternVL3-14B > InternVL3-8B~\citep{zhu2025internvl3} across all three fundamental capabilities. Similar trends are observed for RoboBrain2.0 (32B vs. 7B)~\citep{RoboBrain2.0TechnicalReport} and Qwen2.5-VL-Instruct (7B vs. 3B)~\citep{bai2025qwen2} on embodied QA and embodied task planning. 
Nevertheless, the scaling effect is not consistent: in embodied navigation, smaller models (RoboBrain2.0-7B and Qwen2.5-VL-3B-Instruct) achieve better performance than their larger counterparts.
Moreover, from the perspective of embodied data, increasing the amount of task-specific data can significantly improve models' specific capabilities. For example, StreamVLN~\citep{wei2025streamvln}, NaVILA~\citep{zheng2024towards}, and UniNavid~\citep{zhang2024uni} outperform GPT-o3~\citep{openai2025o3} in instruction-following navigation. However, the constructed embodied datasets usually fail to deliver consistent performance improvements in all capabilities.
For example, Embodied-R1~\citep{yuan2025embodiedr1reinforcedembodiedreasoning} and SpaceR~\citep{ouyang2025spacer}, trained on their respective embodied datasets, surpass the base model Qwen-2.5-VL-3B-Instruct in some capabilities. However, they also suffer from performance drops in others.

\vspace{0.1cm}
\textbf{In-depth Analysis:} 
\emph{The scaling phenomena regarding model size and data amount for embodied models have not emerged generally} across embodied benchmarks and capabilities.
Different from LLMs and general multimodal models, which often share the base model architecture among several canonical choices, embodied models vary considerably in how they are constructed.
Moreover, embodied models are often released in only one or a narrow range of sizes.
Therefore, \emph{a thorough investigation of the scaling law regarding model size for embodied models requires more effort in building consistent model architectures and providing multiple model sizes}.
For the scaling regarding embodied data, \emph{the task-specific performance improvement accompanied by overfitting is likely to stem from insufficient diversity, scope, and scale}.
Hence, a scalable approach for data construction or generation is essential for advancing research in this regard.

\vspace{0.2cm}
\item \textbf{Finding 6:}
\textbf{Reasoning models exhibit strong overall performance on multiple benchmarks by reinforced finetuning (RFT). However, whether RFT can yield stronger out-of-distribution generalization than SFT remains a key open question.}

\vspace{0.1cm}
\textbf{Core Finding:} 
Reasoning models fine-tuned with RFT have demonstrated significant and consistent performance improvements across multiple benchmarks. We observe that the latest records on most benchmarks have been set by these reasoning models. For instance, GPT-o3~\citep{openai2025o3} achieves remarkable and stable performance on task planning benchmarks such as EB-ALFRED, EB-Habitat, and EB-Navigation, exhibiting no obvious capability shortcomings. Space-R~\citep{ouyang2025spacer} establishes a new SOTA for embodied models on OpenEQA~\citep{OpenEQA2023}~(37.70 points), while also maintaining stable performance on other embodied QA tasks. Furthermore, Embodied-R1~\citep{yuan2025embodiedr1reinforcedembodiedreasoning} achieves breakthrough results on affordance prediction tasks like VABench-Point (66 points). In the context of VLN-R1~\citep{qi2025vln}, fine-tuning the Qwen2-VL-7B model~\citep{wang2024qwen2} with VLN data via supervised learning significantly enhances its navigation capabilities. Building upon this, the application of GRPO for RFT further boosts performance, increasing the success rate from 24.9 to 30.2. Nevertheless, although the majority of recently emerged models are reasoning-based (e.g., Gemini-2.5-Pro, o3, and RoboBrain2.0), whether RFT can yield superior generalization compared to SFT and enhance capabilities beyond the training data distribution remains a question that requires and merits further exploration.

\vspace{0.1cm}
\textbf{In-depth Analysis:} 
Lagging behind the development of LLMs and general multimodal models, \emph{the ability of slow thinking or reasoning has not been universally empowered for existing embodied models}.
Pivotal to the ability of slow thinking, RL training has demonstrated its effects in enabling models to activate and combine fundamental perceptual abilities into complex reasoning skills.
It allows the models to \emph{fully utilize fundamental embodied capabilities to address complex tasks, rather than merely pattern matching from training examples}.
This RL-finetuning paradigm is particularly suited for embodied tasks involving multi-step reasoning, sequential decision-making, and precise manipulation, and offers promising directions for future training strategies in embodied AI.

\vspace{0.2cm}
\item \textbf{Finding 7:}
\textbf{3D representations are essential for embodied understanding but face challenges in the alignment with language modality. Strategic integration with 2D-3D representation can effectively leverage pre-trained language alignment to unlock superior spatial understanding.}

\vspace{0.1cm}
\textbf{Core Finding:} The evaluation results on 3D Embodied QA benchmarks show that models that effectively integrate 2D visual features with 3D spatial priors significantly outperform those relying solely on naive 3D data processing. The top-performing models --- GPT4Scene-HDM~\citep{qi2025gpt4scene} (71.00), LL3DA~\citep{chen2024grounded} (62.11), 3DRS (65.77), OmniEVA (64.66), and Video-3D LLM~\citep{zheng2025video} (64.92) --- all adopt visual-spatial integration strategies, while the models that adopt native 3D processing methods like  LEO~\citep{huang2023embodied} (48.48) consistently underperform with a 15-25\% performance gap.

\vspace{0.1cm}
\textbf{In-depth Analysis:} 
The evaluation reveals a fundamental architectural principle: \emph{3D geometric representations provide important spatial awareness that 2D understanding cannot deliver}, yet they encounter  challenges in achieving sufficient language modality alignment.
Traditional approaches that directly process point clouds or voxels through 3D encoders show consistently poor performance across 3D embodied benchmarks.
This indicates that
\emph{naive 3D representation methods face inherent challenges in language modality alignment and cannot effectively leverage pre-training model capacity}. 
In contrast, current leading models consistently employ strategic integration methods that \emph{combine rich 2D visual features with explicit 3D spatial information through various encoding mechanisms} through position encoding, multi-view synthesis, coordinate injection, etc.
The above represents a compromise technical solution constrained by the current lack of native 3D foundation language models. However, from a long-term perspective, exploring \emph{how to achieve in-depth alignment between native 3D information and language} through multi-stage training mechanisms or innovative architectural designs remains a more critical research direction in the field of embodiment.

\vspace{0.2cm}

\item \textbf{Finding 8:} \textbf{Embodied Navigation methods can be derived by harnessing models with either end-to-end or agentic framework: E2E frameworks show VLN-specialized models outperforming general models through enhanced embodied capabilities, while agentic frameworks achieve better performance via structured pipeline design to integrate extensible modular architecture and external knowledge especially for long-horizon tasks.}

\vspace{0.1cm}
\textbf{Core Finding:} E2E frameworks show that VLN-specialized models demonstrate substantial performance advantages over general multimodal models through targeted architectural innovations and domain-specific training data. The top-performing VLN models --- StreamVLN~\citep{wei2025streamvln} (54.90\%), NaVILA~\citep{zheng2024towards} (51.65\%), and UniNavid~\citep{zhang2024uni} (47.85\%) --- achieve dramatically higher success rates compared to leading general models like Claude-3.7-Sonnet~\citep{claude37sonnet2024} (20.17\%), Gemini-2.5-Pro~\citep{gemini2024flash} (27.37\%), and GPT-4o~\citep{openai2024gpt4o} (21.67\%). Notably, specialized VLN models dominate the entire top-5 rankings, with even mid-tier VLN models like MapNav~\citep{zhang2025mapnav} (36.15\%) and Navid~\citep{zhang2024navid} (30.60\%) outperforming most general foundation models. This performance gap is particularly pronounced in navigation-specific metrics, where StreamVLN~\citep{wei2025streamvln} achieves 56.90\% on VLN-CE R2R compared to general models that struggle to exceed 25\% success rates. 
And agentic frameworks demonstrate the potential to alleviate these built-in model limitations: specialized agentic frameworks like OmniEVA achieve top performance (59.10\% on MP3D, 74.20\% on HM3D) and OVRL~\citep{yadav2023offline} also achieve competitive results (62.00\% on HM3D).

\vspace{0.1cm}
\textbf{In-depth Analysis:} This performance gap between VLN-specialized and general models stems from architectural and training differences addressing embodied navigation challenges. \emph{VLN-specific architectures demonstrate superior performance by leveraging historical frames rather than relying solely on current frames}, e.g., models like NaVid~\citep{zhang2024navid} and UniNavid~\citep{zhang2024uni} allocate more tokens to current frames for improved decision accuracy. 
The combination of RxR and R2R datasets with Habitat simulation enables \emph{large-scale vision-language-action data construction for VLN}, allowing effective supervised fine-tuning and strong validation performance on unseen splits.
Building upon these capabilities, \emph{designing efficient hierarchical agentic frameworks} represents a promising direction for leveraging VLN capabilities. Such frameworks could decompose complex navigation tasks into subtasks, integrate multi-modal reasoning with spatial planning, and provide error recovery mechanisms. Agentic approaches \emph{activate and amplify existing foundation model capabilities through external reasoning pipelines rather than requiring costly model retraining or architectural modifications}. This enables consistent performance enhancement through pipeline optimization, where new knowledge sources, memory architectures, and reasoning strategies can be systematically integrated without modifying the foundation model itself.

\vspace{0.2cm}
\item \textbf{Finding 9:} \textbf{Embodied pointing is critical to both enhancing fundamental embodied capabilities and improving downstream embodied task performance.
Supervised/reinforced fine-tuning for pointing not only significantly enhances performance on pointing tasks, but can also lead to improvements of fundamental embodied abilities. Pointing tasks under complex instructions remain a major challenge for most models.}

\vspace{0.1cm}
\textbf{Core Findings:} We evaluate the pointing capabilities of models on Where2Place~\citep{yuan2024robopoint} and VABench-P~\citep{yuan2025seeingdoingbridgingreasoning}. The data shows that training on dedicated pointing data significantly boosts pointing performance. On the Where2Place benchmark, the top three performers---RoboBrain2.0~\citep{RoboBrain2.0TechnicalReport}, Embodied-R1~\citep{yuan2025embodiedr1reinforcedembodiedreasoning}, and RoboRefer~\citep{zhou2025roborefer} are all embodied models fine-tuned with pointing data. For instance, RoboBrain2.0-7B improved by 45.65\% after incorporating data from the Spatial Referring Dataset~\citep{zhou2025roborefer}. However, performance diverges sharply on the more challenging VABench-P benchmark. While most top-tier models score above 60 on Where2Place, they fail to surpass 40 on VABench-P, with only Embodied-R1 (66) and Qwen-VL-Max (42) as notable exceptions. Moreover, models specifically optimized for pointing, such as RoboPoint (19.09) and Roborefer (4.62), also underperform significantly on this benchmark. More importantly, enhancing pointing capability appears to promote the model's generalization on other tasks. Taking Embodied-R1 as an example, after RFT on a dataset containing partial spatial reasoning and pointing data, it achieved stable performance improvements across several OOD benchmarks, including OpenEQA (26.19 $\rightarrow$ 34.51), ERQA (32.61 $\rightarrow$ 35.24), and UniEQA (33.62 $\rightarrow$ 38.14).

\vspace{0.1cm}
\textbf{In-depth Analysis:}  
Why do models optimized for pointing perform well on Where2Place but exhibit uncertainty on VABench-P? We posit that this discrepancy is primarily attributed to the inherent complexity of VABench-P. Compared to Where2Place, its tasks feature more intricate instructions and diverse scenes, requiring models to seamlessly integrate instruction understanding, spatial reasoning, and multimodal pointing capabilities. This integrated challenge reveals a critical trade-off: \emph{some embodied models, despite being fine-tuned on pointing data, appear to overfit. This specialization on specific pointing tasks may weaken their broader understanding and reasoning abilities, causing poor performance when faced with varied instructions or novel tasks}. Fundamentally, the enhancement of general capabilities through pointing training is rooted in its role as a critical \textit{grounding mechanism} for embodied AI. It compels the model to anchor abstract language to precise spatial coordinates, thereby serializing and integrating sub-tasks like perception, reasoning, and planning onto points. This anchoring process strengthens the model's cognitive integration and boosts its generalization capabilities, as evidenced by Embodied-R1's strong performance on several OOD benchmarks. Therefore, pointing tasks in complex environments are not merely about localization; they are intuitive, expressive, and provide the precise anchor points required for subsequent manipulation, making them an effective metric for evaluating multimodal understanding and reasoning~\citep{cheng2025pointarena}. In summary, \emph{mastering embodied pointing remains a crucial core capability that advanced embodied models must develop}.

\end{itemize}

\vspace{0.2cm}
For complete and detailed discussions of the evaluation results on each benchmark, please refer to the leaderboard page of Embodied Arena website\footnote{Please refer to the detailed discussions below the leaderboard tables in the website \url{https://embodied-arena.com}.}.

\section{Conclusion}
\label{sec:conclusion}

We introduce Embodied Arena, a comprehensive, unified, evolving evaluation platform and leaderboards for embodied AI models. It features three types of core embodied tasks, a diverse range of high-quality benchmarks, an LLM-driven automated evaluation data generation framework, and a systematic embodied capability taxonomy. Moreover, Embodied Arena offers professional support for advanced models and new benchmarks to join. 
With three types of real-time leaderboards and two evaluation views, Embodied Arena presents a multifaceted overview of embodied capabilities of advanced models.
This offers a convenient way for researchers in both academia and industry to obtain useful insights and helps pinpoint critical research directions, thereby propelling the research progress in the field of Embodied AI. As the field evolves toward more sophisticated embodied agents, future extensions of the platform will incorporate more comprehensive manipulation tasks and closed-loop evaluation capabilities.

\section{Contributions}
\label{sec:contribution}

This work represents a collaborative effort from researchers across multiple institutions worldwide. 
The authors are listed below:

\textbf{Contributors:} Fei Ni,  Min Zhang, Pengyi Li, Yifu Yuan, Lingfeng Zhang, Yuecheng Liu, Peilong Han, Longxin Kou, Shaojin Ma, Jinbin Qiao, David Gamaliel Arcos Bravo, Yuening Wang, Xiao Hu, Zhanguang Zhang, Xianze Yao, Yutong Li, Zhao Zhang, Ying Wen, Ying-Cong Chen, Xiaodan Liang, Liang Lin, Bin He, Haitham Bou-Ammar, He Wang, Huazhe Xu, Jiankang Deng, Shan Luo, Shuqiang Jiang, Wei Pan, Yang Gao, Stefanos Zafeiriou, Jan Peters, Yuzheng Zhuang, Yingxue Zhang, Yan Zheng, Hongyao Tang, Jianye Hao.

The development of Embodied Arena involved contributions across multiple areas including benchmark integration, model evaluation infrastructure, automated data generation pipelines, capability taxonomy design, and comprehensive analysis. Each contributor brought expertise from their respective institutions to create this unified evaluation platform for Embodied AI.

\bibliography{main}

\end{document}